\documentclass{article}

\usepackage{arxiv}

\usepackage[utf8]{inputenc} 
\usepackage[T1]{fontenc}    
\usepackage{hyperref}       
\usepackage{url}            
\usepackage{booktabs}       
\usepackage{amsfonts}       
\usepackage{nicefrac}       
\usepackage{microtype}      
\usepackage{lipsum}		
\usepackage{graphicx}
\usepackage{natbib}
\usepackage{doi}

\usepackage{amssymb}
\usepackage{amsthm}
\usepackage{amsmath}
\usepackage{caption}
\usepackage{subcaption}

\makeatletter
\newcommand{\printfnsymbol}[1]{%
  \textsuperscript{\@fnsymbol{#1}}%
}
\makeatother

\title{On Disentanglement in Gaussian Process Variational Autoencoders}

\date{} 					

\author{Simon Bing\thanks{Equal contribution.}\hspace{1.5mm}\thanks{\texttt{bings@ethz.ch}} \\
	Department of Computer Science\\
	ETH Z\"urich\\
	Z\"urich, Switzerland \\
	\And
	Vincent Fortuin\printfnsymbol{1} \\
	Department of Computer Science\\
	ETH Z\"urich\\
	Z\"urich, Switzerland \\
	\And
	Gunnar R\"atsch \\
	Department of Computer Science\\
	ETH Z\"urich\\
	Z\"urich, Switzerland \\
}

\renewcommand{\headeright}{Preprint}
\renewcommand{\undertitle}{}

\hypersetup{
pdftitle={On Disentanglement in Gaussian Process Variational Autoencoders},
pdfsubject={stat.ML, cs.LG},
pdfauthor={Simon~Bing, Vincent~Fortuin, Gunnar~R\"atsch},
pdfkeywords={Disentanglement, Time Series, Gaussian Process, Machine Learning},
}

\begin{document}
\maketitle

\begin{abstract}
Complex multivariate time series arise in many fields, ranging from computer vision to robotics or medicine. Often we are interested in the independent underlying factors that give rise to the high-dimensional data we are observing.
While many models have been introduced to learn such \emph{disentangled} representations, only few attempt to explicitly exploit the structure of sequential data. We investigate the disentanglement properties of Gaussian process variational autoencoders, a class of models recently introduced that have been successful in different tasks on time series data.
Our model exploits the temporal structure of the data by modeling each latent channel with a GP prior and employing a structured variational distribution that can capture dependencies in time.
We demonstrate the competitiveness of our approach against state-of-the-art unsupervised and weakly-supervised disentanglement methods on a benchmark task. Moreover, we provide evidence that we can learn meaningful disentangled representations on real-world medical time series data.
\end{abstract}


\section{Introduction}
\label{intro}

\begin{figure}[t]
    \centering
    \includegraphics[width=0.45\textwidth]{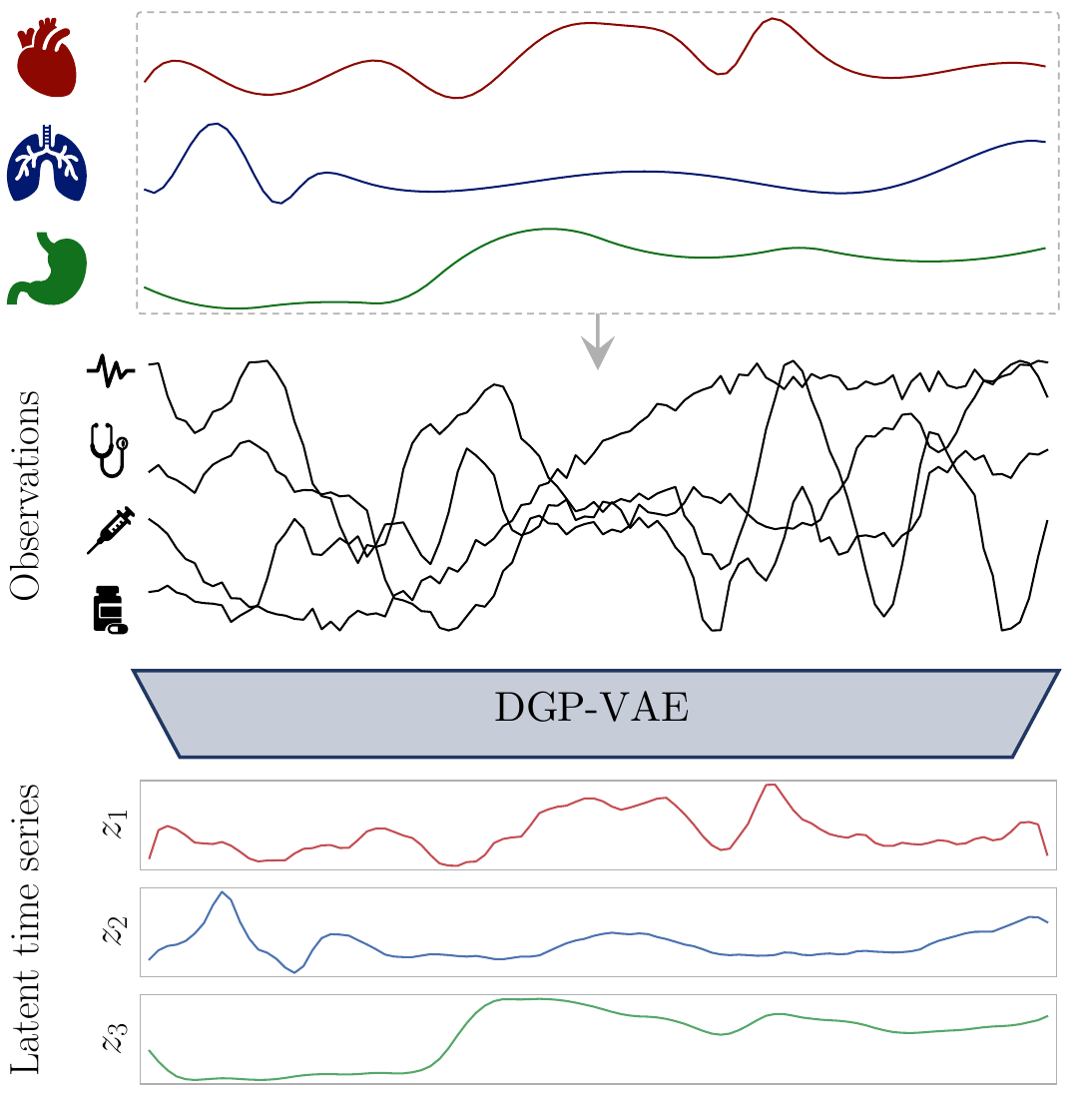}
    \caption{Conceptual overview of an application of our model in a medical setting. The underlying physiological states of different organ systems (\textbf{top}) give rise to interdependent, multivariate observational time series (\textbf{middle}). Our model learns a disentangled representation from these sequential observations and provides smooth latent time series (\textbf{bottom}), that are more interpretable than the noisy, high-dimensional observations.}
    \label{fig:overview}
\end{figure}

The success of any machine learning application is greatly influenced by the representation of the provided data. \citet{bengio1} state that if a representation caters to a certain learning task, better results as well as increased robustness may be expected. Data representations found and used in recent successful deep learning approaches for certain tasks \citep{go,He2016DeepRL,Mnih2015HumanlevelCT} overfit to the task at hand \citep{annealedvae}, which partially explains why they still fall behind biological intelligence in terms of generality and knowledge transfer \citep{lake2016building}. This motivates the learning of task-agnostic representations that fare well on a wide variety of problems. Recent research suggests that disentangled representations could be suitable in this sense \citep{bengio1,ridgeway2016survey,tschannen2018recent}. 

While there is no single, widely accepted definition of disentanglement in the literature, the intuition of what constitutes such a representation is shared. Disentangled representations should separate the distinct, independent factors of variation that led to the generation of said data \citep{bengio1}. Such disentangled representations have the property that a single latent factor is only sensitive to a change in a single underlying factor of variation, while being invariant under changes to any other factors of variation. It has been argued that such representations are akin to how humans learn and transfer knowledge, eventually leading to possible advances in settings where human intelligence currently exceeds that of machines \citep{betavae}. Moreover, disentangled representations have been argued to offer benefits in terms of interpretability \citep{adel18a,bengio1}, predictive performance \citep{challengingdis}, fairness \citep{Locatelloetal19b} and reducing the sample complexity for downstream tasks \citep{steenkiste}.

Sequential data appears in a wide variety of settings such as audio and video streams, communication signal processing, or longitudinal medical data, motivating the investigation of learning disentangled representations from such data. While there has been previous work on disentangling static from dynamic factors of sequential data \citep{seqdvae,fhvae}, none have attempted to disentangle the dynamic factors themselves. Previously introduced models belonging to the class of Gaussian process variational autoencoders (GP-VAEs) have very successfully leveraged the temporal correlations of such sequential data to tackle problems such as conditional generation \citep{gppvae} or missing value imputation \citep{gpvae}, but have not explicitly looked at disentangling such data. 

Recent work in the disentanglement literature has shown that learning disentangled representations in a fully unsupervised fashion is fundamentally impossible \citep{challengingdis}. \citet{adagvae} show that inductive biases must be included to achieve this task and further provide the first model that makes explicit assumptions on the structure of the input data to improve the disentanglement of the learned representation. Inspired by the notion of weak supervision in this model as well as the the successful application of GP-VAE models to tasks involving sequential data, we investigate the disentanglement properties of GP-VAE type models. To this end we study the Disentangled Gaussian Process Variational Autoencoder (DGP-VAE), an adaptation of \citet{gpvae}'s GP-VAE model, that exploits the sequential structure of time series data to learn disentangled representations.

We make the following contributions:
\begin{itemize}
    \item We study the disentanglement properties of the recently proposed GP-VAE model, by introducing a modification, the DGP-VAE, where latent GP priors with variable length scales are used to encourage disentanglement between latent dimensions.
    \item We compare against state-of-the-art disentanglement models and show that we outperform all considered baselines in terms of disentanglement on standard benchmark data sets. 
    \item We perform a study on real-world medical time series data and demonstrate that our modeling assumptions are better suited to learning disentangled representations of real time series data compared to those of previously introduced weakly-supervised models.
\end{itemize}

 Our paper starts by expanding on the specific problem setting in Sec.~\ref{prob_setting}, followed by the introduction of our generative model in Sec.~\ref{gen_model} and inference model in Sec.~\ref{inf_model}. In Sec.~\ref{related} we review the related literature on VAEs and disentanglement. We present our two  main experiments and their results in Sec.~\ref{base_methods} and Sec.~\ref{real_med}, respectively. Finally, we present our conclusions in Sec.~\ref{conclusion}.

\section{Disentangled Gaussian Process VAE}
\label{model}
We introduce the disentangled Gaussian process variational autoencoder (DGP-VAE), a GP-VAE-based model \citep{gpvae} for learning disentangled representations from time series data. The main idea of our model is to exploit the correlations in sequential data sets by the application of Gaussian process latent priors. We make certain smoothness assumptions about the input data and explicitly exploit this inductive bias for the benefit of disentanglement. We argue that while our assumptions on the dynamics of the sequential data are weaker than previous approaches, they are better aligned with real-world data. We provide evidence for this claim on a data set consisting of real-world medical time series in Sec.~\ref{real_med}.

In the following, we will outline the general setting of disentangled representations in VAEs, introduce the notation and the problem setting, describe the employed generative model and formulate the inference model with its specific inductive bias for sequential data.

\subsection{VAEs and Disentanglement}
\label{vaes_disent}
As we state in Sec.~\ref{related}, all state-of-the-art approaches to disentangled representation learning share VAEs as the basis of their architecture. In this setting, \(p(\textbf{z})\) denotes the prior distribution assumed for the latent space. The conditional distribution of a sample \(\textbf{x}\) given the latent code \(\textbf{z}\), \(p(\textbf{x}|\textbf{z})\), is parameterized by a deep neural network---commonly referred to as the decoder network---which gives rise to the notation \(p_{\theta}(\textbf{x}|\textbf{z})\). Since the posterior distribution \(p(\textbf{z}|\textbf{x})\) is intractable to compute, it is approximated by a variational distribution \(q_{\psi}(\textbf{z}|\textbf{x})\), which is again parameterized by a deep neural network referred to as the encoder network. Performing optimization on the marginal log-likelihood is intractable, so both networks are trained by optimizing its lower bound,
\begin{equation*}
    \max_{\psi,\theta} \mathbb{E}_{p(\textbf{x})} 
    \left[ 
    \mathbb{E}_{q_{\psi}(\textbf{z}|\textbf{x})}[
    \log p_{\theta}(\textbf{x}|\textbf{z})] 
    - D_{\text{KL}}(q_{\theta}(\textbf{z}|\textbf{x})||p(\textbf{z})) 
    \right],
\end{equation*}
commonly referred to as the evidence lower bound (ELBO).

The representation \(r(\textbf{x})\), that maps an observation to the latent space, is then given by the encoder network, parameterized by \(\psi\). All considered baseline models, as well as ours, leverage the influence of the KL-term in the ELBO to encourage a more disentangled latent representation.

\subsection{Notation and Problem Setting}
\label{prob_setting}
As we are considering learning disentangled representations from sequential data, we assume that one sample \(\textbf{x}_{1:T} \in \mathbb{R}^{T \times d}\) consists of \(T\) steps \(\textbf{x}_t = [x_{t,1},x_{t,2},...,x_{t,d}]^\top \in \mathbb{R}^d\). Furthermore, we assume that each step corresponds to a consecutive sample from a temporally coherent time series, that is, we explicitly do \textit{not} assume that each sample is i.i.d. We assume that each \(\textbf{x}_t\) comes from some underlying generative process where \(k\) generative factors of variation fully determine the outcome. These factors are independent across dimensions, but not in time, meaning that for each time series \(\textbf{x}_{1:T} \in \mathbb{R}^{T \times d}\) there is also an underlying generative time series \(\textbf{c}_{1:T} \in \mathbb{R}^{T \times k}\). In general, it holds that \(d \geq k\). In contrast to \citet{adagvae}, we do not make the assumption that the change in the underlying generative time series must be sparse nor that a certain number of factors must be shared from one time step to the next. Rather, we allow for the more general setting of dense changes in the underlying factors from one time step to the next.  It should be noted that we never observe the generative factors \textbf{c}.

Given this setting, we aim to learn a representation function \(r(\textbf{x}_{1:T})\) that takes as input a time series \(\textbf{x}_{1:T}\) and maps it to a representation \(\textbf{z}_{1:T} \in \mathbb{R}^{T \times m}\), where \(m\) is the resulting latent dimensionality. The goal is hereby to find a latent representation \(\textbf{z}_t = [z_{t,1},z_{t,2},...,z_{t,m}]^\top \in \mathbb{R}^m\) that is as \textit{disentangled} as possible. The exact measure used to determine the degree of disentanglement is described in Sec.~\ref{metric}, but intuitively the goal is that dimensions in the learned $\textbf{z}_t$ should correspond as much as possible to independent dimensions in the true underlying $\textbf{c}_t$.

\subsection{Generative Model}
\label{gen_model}
Previous work has shown that learning fully disentangled representations is fundamentally impossible without inductive biases and since they are necessary for the task at hand, it is beneficial to be explicit about these modeling assumptions \citep{challengingdis}. This allows one to leverage the assumptions to improve the disentanglement properties of the learned representations. Following this impossibility result, there have been approaches that break with the paradigm of the fully unsupervised setting \citep{fewlabels} and no longer assume the input data to be i.i.d. \citep{adagvae}. This setting naturally lends itself to data occurring in time series. While these models have been shown to outperform their fully unsupervised counterparts in terms of disentanglement, we argue that their assumptions that a certain number of underlying generative factors must be shared from one time step to the next are too restrictive for realistically occurring time series data.

We provide a less restrictive set of assumptions on sequential data, namely that data occurring in a given sequence is not i.i.d., that is, there is a certain degree of temporal coherence as well as that the change of the data in time is smooth to some degree. Therefore, the time series of the latent representations is also assumed to be smooth and correlated in time.  We model these assumptions in the form of a GP prior, \(\textbf{z}(\tau) \sim \mathcal{GP}(m_{\textbf{z}}(\cdot), k_{\textbf{z}}(\cdot,\cdot))\), for each latent channel. Gaussian processes have been previously shown to be suitable for time series modeling \citep{Roberts2013GaussianPF}, and we further justify this choice by the ability to model correlations of points in time while enforcing smoothness, which is congruent with the aforementioned assumptions we make. Since each latent channel is modeled with its own GP prior, these remain independent. 

This choice of latent prior distribution further grants our model the additional flexibility to capture a plethora of possible temporal coherence characteristics in the data, by employing different GP kernels. The Ada-GVAE model \citep{adagvae} can be seen as a special case of this architecture, albeit with a very rigid kernel that only captures pairwise correlations, as opposed to our more flexible architecture that can capture long-range dependencies. As multivariate time series often exhibit dynamics on a number of different timescales, we opt for the Cauchy kernel in practice
\begin{equation}
    k_{Cau}(\tau, \tau') = \sigma^2 \left(1 + \frac{(\tau - \tau')^2}{l_i^2}\right)^{-1},
    \label{eq:cauchy}
\end{equation}

which can be derived from an infinite mixture of radial basis function (RBF) kernels. The Cauchy kernel therefore naturally lends itself to modelling dynamics on a variety of time scales. This choice of kernel can be viewed as generally applicable, but more specific kernels could also be employed, given prior knowledge of the data in question. We have chosen not to further investigate this, as the Cauchy kernel yielded satisfactory results in all of our applications.
Note that in our DGP-VAE model, in contrast to the original GP-VAE \citep{gpvae}, the length scale $l_i$ is variable between different latent channels, in order to encourage the disentanglement of factors of variation that change with different frequencies over time.

The conditional distribution of an observation \(\textbf{x}_t\), given the latent sequence \(\textbf{z}_{1:T}\), is modeled point-wise as
\begin{equation}
    p_\theta(\textbf{x}_t|\textbf{z}_t) = \mathcal{N}(g_\theta(\textbf{z}_t),\sigma^2 \mathbf{I}),
    \label{eq:gen_dist}
\end{equation}

where the possibly nonlinear function \(g_\theta(\textbf{z}_t)\) is approximated by a deep neural network.

\subsection{Inference Model}
\label{inf_model}
The inference model in our architecture yields the approximate posterior distribution \(q_{\psi}(\textbf{z}_{1:T}|\textbf{x}_{1:T})\), which is needed to infer the latent representation from the input data and to learn the parameters of the previously introduced generative model. 
Since inferring the exact posterior distribution is intractable, we employ a variational inference scheme \citep{Jordan99,Blei17}. Based on the architecture proposed by \citet{gpvae}, we use a structured variational distribution \citep{MAL-001} to capture the temporal correlation of the data, in conjunction with efficient amortized inference \citep{vae}.

The family of variational distributions that approximates the posterior is chosen to be a Gauss-Markov process in the time domain. Specifically, the approximate posterior distribution for a given latent dimension \(j\) is chosen to be a multivariate Gaussian variational distribution
\begin{equation}
    q(\textbf{z}_{1:T,j}|\textbf{x}_{1:T}) = \mathcal{N}(\textbf{m}_j,\boldsymbol{\Lambda}^{-1}_j),
\end{equation}
where the structure of the precision matrix \(\boldsymbol{\Lambda}_j\) is achieved by means of the parameterization
\begin{equation}
    \begin{split}
    & \boldsymbol{\Lambda}_j := \textbf{B}^{\top}_j \textbf{B}_j, \\ & \text{with } \{ \textbf{B}_j \}_{tt'} = 
    \begin{cases}
    b^j_{tt'} & \text{if } t' \in \{t,t+1\} \\
    0 & \text{otherwise}
    \end{cases}.
    \end{split}
\end{equation}
\(\textbf{B}_j\) is an upper triangular band matrix and \(b^j_{tt'}\) are the variational parameters. This choice of structure for the variational distribution, reminiscent of that of \citet{Bamler17}, allows our model to capture dependencies in time, while breaking them across latent dimensions. This facilitates the effective leveraging of the assumed temporal correlation of data points, while maintaining the inter-dimensional independence we require for disentanglement. While the precision matrix \(\boldsymbol{\Lambda}_j\) is sparse, the resulting covariance matrix does not necessarily have to be, allowing our model to capture long-term correlations in time. Furthermore, this formulation results in the approximate posterior being a GP itself.

We jointly train the generative network parameters \(\theta\) and the inference network parameters \(\psi\) by optimizing the following objective
\begin{equation}
\begin{split}
     \max_{\psi, \theta} \sum_{t=1}^{T} \; &\mathbb{E}_{q_\psi(\textbf{z}_t|\textbf{x}_{1:T})}[\log p_\theta(\textbf{x}_t|\textbf{z}_t)] \\ 
     & - \beta D_{\text{KL}}(q_\psi(\textbf{z}_{1:T}|\textbf{x}_{1:T})||p(\textbf{z}_{1:T})),
    \end{split}
    \label{eq:gpvaeobj}
\end{equation}
where \(\beta\) is a hyperparameter used to balance reconstruction quality with latent channel capacity, as first introduced in the \(\beta\)-VAE model \citep{betavae}.

Our objective function reflects the fact that our model can be seen as an extension of previously successful disentanglement methods for sequential data. To this end, we use the GP-VAE framework with a suitable prior distribution for time series data in conjunction with a structured variational distribution that is well suited to model temporal correlations.

In summary, our model can be seen as a modified version of the GP-VAE model from \citet{gpvae} in which we have a higher fidelity of control over the independent GP priors of the different latent dimensions, in order to encourage the disentanglement of underlying dynamic factors which vary with different rates.
While this is not a major extension of the model, our goal in this work is not to propose a radically new model, but to study the general disentanglement properties of GP-VAE models when imbued with reasonable assumptions about realistic time series data.
In Sec.~\ref{sec:experiments} we will show how we leverage our modeling assumptions to learn disentangled representations from real-world time series data, and show competitive results against state-of-the-art disentanglement methods.

\section{Related Work}
\label{related}

In the following, we give a brief overview over related work in this field.
A more detailed exposition is deferred to Appendix~\ref{sec:detailed_related}.

\begin{figure*}[t]
    \centering
    \begin{subfigure}{.9\textwidth}
    \includegraphics[width=\linewidth]{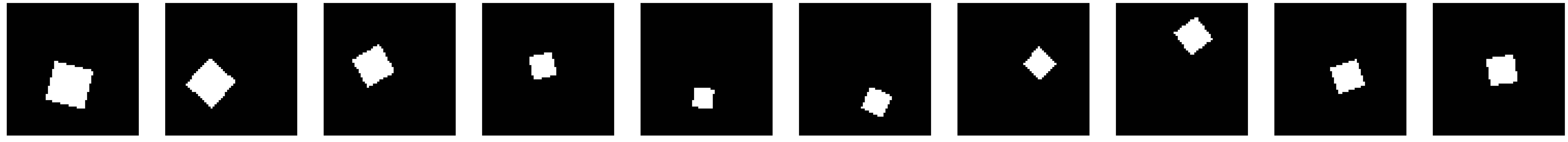}  
    \end{subfigure}
    \begin{subfigure}{.9\textwidth}
    \includegraphics[width=\linewidth]{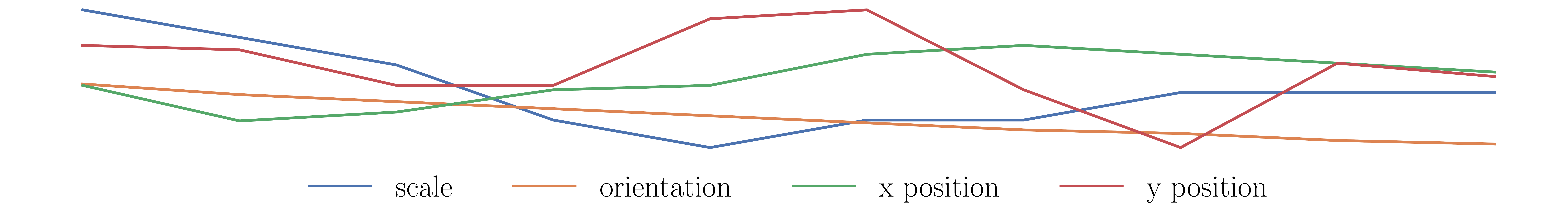}  
    \end{subfigure}
    \caption{Underlying generative time series (\textbf{bottom}) and the resulting sampled time series for the dSprites data set (\textbf{top}).}
    \label{fig:data_gen}
\end{figure*}

\paragraph{Sequential Data in VAEs}
Previous approaches to apply VAEs to sequential data include the factorized hierarchical VAE (FHVAE) \citep{fhvae}, which aims to exploit the correlations of sequential data by introducing two different hierarchical priors into the latent representation, and the Disentangled Sequential Autoencoder \citep{seqdvae}, a model that learns to disentangle static from dynamic parts of the data in the representation. While both of the aforementioned models successfully disentangle static from dynamic features, they do not disentangle the underlying, \textit{individual} dynamic factors, in contrast to our model.

\paragraph{Gaussian Process VAEs}
Gaussian process VAEs (GP-VAEs) employ Gaussian process priors in the latent space of a VAE, thereby explicitly modeling correlations between data points, which in principle enables them to model correlations over time.
The Gaussian Process Prior Variational Autoencoder (GPPVAE) \citep{gppvae} was the first model to introduce this family of priors in the context of VAEs. It was however not specifically designed for sequential data, but for conditional generation of images with different types of auxiliary data.
The GPPVAE was extended by the GP-VAE \citep{gpvae}, which explicitly takes sequential data into account.
While the GP-VAE was designed and used for missing data imputation, we show that a very similar architecture (our DGP-VAE) lends itself to learning disentangled representations from dynamic sequential data.
Another extension of the GPPVAE is the Multi-disentangled-features Gaussian Process Variational Autoencoder (MGP-VAE) \citep{mgpvae}, which uses fractional Brownian motion and Brownian bridge kernels for the latent GP prior of each channel. The authors argue that this setting allows for the disentanglement of static as well as dynamic features, but only show qualitative results for sparsely changing input time series. In contrast, we show the efficacy of our approach to dynamic sequential data with dense changes of the factors in time.
This general class of models has been further extended by the Sparse GP-VAE (SGP-VAE) \citep{sgpvae}, the Scalable GP-VAE (SVGP-VAE) \citep{jazbec2020scalable}, and the Factorized GP-VAE (FGP-VAE) \citep{jazbec2020factorized}, which all improve their scalability to larger data sets.
These extensions could also be readily applied to our model, which however we have not done in this study, since exact inference was still feasible in our experiments.

\paragraph{Unsupervised Disentanglement Models}
All state-of-the-art approaches for disentangled representation learning rely on variational autoencoders \citep{vae} as their architectural backbone. Alternatives based on generative adversarial networks (GANs) \citep{gan} such as variants of InfoGAN \citep{infogan} have also been proposed, but previous work has found their performance to not be competitive when compared to VAE-based approaches \citep{factorvae}.
State-of-the-art VAE methods that encourage disentanglement in the latent representations through additional regularization of the optimization objective include the \(\beta\)-VAE \citep{betavae}, the AnnealedVAE \citep{annealedvae}, the \(\beta\)-TCVAE \citep{betatcvae}, the FactorVAE \citep{factorvae}, and the DIP-VAE \citep{dipvae} (type I and II).
However, none of these explicitly take sequential data into account and we show empirically that our DGP-VAE outperforms all of them on time series data.

\paragraph{Weakly-Supervised Disentanglement Model}
The Ada-GVAE model \citep{adagvae} differs from the previously introduced approaches in that it utilizes a form of weak supervision to improve disentanglement. The authors acknowledge \citet{challengingdis}'s proof that learning disentangled representations is impossible without inductive biases. They therefore explicitly make the assumption that some underlying factors of variation are shared across pairs of input data.
The authors argue that a source of data where this assumption could be justified is sequential data, thus making this model the main competitor for our DGP-VAE approach.
Instead of their strong factor-sharing assumption, we make weaker smoothness assumptions which we incorporate into the kernel of our GP priors.
We believe that our assumptions are more aligned with real-world time series and provide evidence for this in the experiments.

\section{Experiments}
\label{sec:experiments}


We performed experiments on time series data synthesized from four different data sets commonly used in the disentanglement literature: dSprites \citep{dsprites17}, SmallNORB \citep{norb}, Cars3D \citep{cars3d} and Shapes3D \citep{3dshapes18}. We provide an extensive comparison against state-of-the-art unsupervised approaches \citep{betavae,annealedvae,betatcvae,factorvae,dipvae} and a recently proposed state-of-the-art weakly-supervised model \citep{adagvae} for disentangled representation learning. 
We present strong quantitative evidence that our model outperforms all of the competing approaches in learning disentangled representations from sequential data.

Additionally, we investigated the disentanglement properties of our model when applied to real-world medical time series data. To this end, we performed an experiment using the HiRID data set \citep{hirid} and compare against the weakly-supervised Ada-GVAE model \citep{adagvae}. This setting allows us to validate our model's applicability to real-world time series data, while demonstrating that our modeling assumptions are more closely aligned with such data than the more restrictive assumptions made by the Ada-GVAE.

In the following, we will first introduce the experimental setup of the benchmark experiments, including an overview of the data sets considered and how we synthesize time series from them, followed by an introduction of the metric we use to measure disentanglement. We then present the experimental findings of both aforementioned experiments.
Implementation details for all experiments can be found in Appendix~\ref{app:imp_details} and additional results in Appendix~\ref{app:add_results}.
All of our code is available online\footnote{\url{https://github.com/ratschlab/dgp-vae}}.

\subsection{Standard Benchmark Data}
\label{base_methods}

\paragraph{Experimental Setup}
To facilitate a fair comparison against state-of-the-art disentanglement approaches, we synthesize sequential data from four publicly available and commonly used data sets to investigate disentanglement. These synthetic data sets lend themselves to our experimental purposes for the practical reason that we have access to the underlying factors of variation for each observed sample. We need these underlying factors to synthesize time series of observations, as well as to qualitatively evaluate the disentanglement of the learned representations.

To create sequential data sets from the existing benchmark data, we first sample a time series for each underlying factor of variation that characterizes the respective data.
These time series are generated by sampling from a GP with a kernel consisting of a mixture of an RBF and a constant value kernel. Different underlying generative factors are independent and vary with different time scales, by varying the length scale of the RBF component of each channel's GP kernel.
We then use this generative factor time series to generate time series of observations point-wise in time. This generative process is illustrated in \textbf{Fig.}~\ref{fig:data_gen} and is conceptually identical for all considered data sets.

In contrast to \citet{adagvae}, we do not impose any restriction on the number of underlying factors that change from one time step to the next and in general this change can be dense, that is, all underlying factors may change. 
We believe that this reflects the nature of multivariate time series in the real world, such as medical time series, where the underlying causes of observed variables may exhibit dynamics on a wide range of time scales.
For instance, a patient's blood pressure will change much more rapidly than their blood sugar or liver hormone levels, but all of them may change simultaneously at any given time.

To ensure that we do not overfit to a single setting, we consider a variety of data sets, each with a different number of underlying generative factors. Using the previously introduced generative process, we synthesize time series consisting of 100 samples each. For the dSprites and SmallNORB data sets, 10,500 such time series are sampled and for the Cars3D and Shapes3D data sets, 6,500 time series.
The considered baseline methods\footnote{The baseline methods considered are implemented with the \texttt{disentanglement\_lib} \citep{challengingdis}.} and our model are all trained on the same data sets. The pairs required for the training of the Ada-GVAE model are taken as neighboring samples within a time series, as suggested in the original paper \citep{adagvae}.

\begin{figure*}
\centering
    \includegraphics[width=\textwidth]{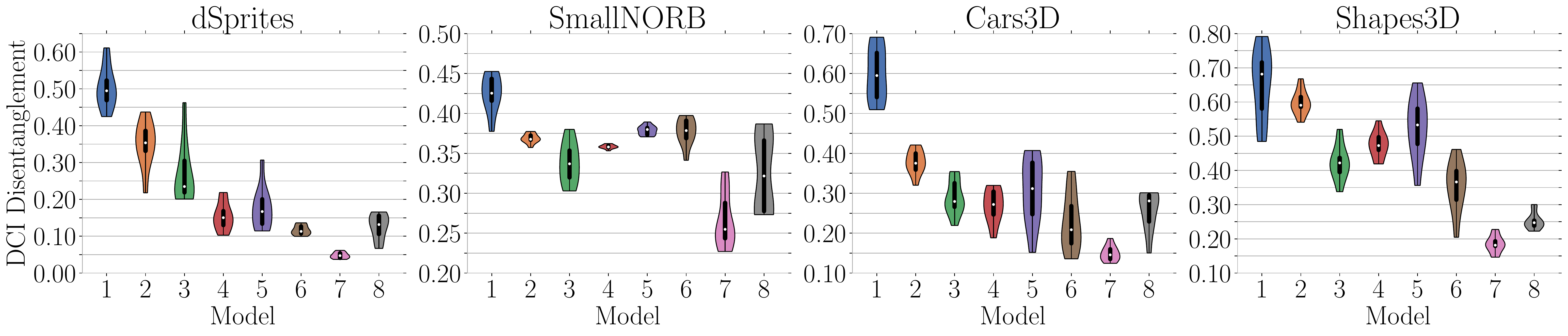}
    \caption{Results of the baseline experiments for all considered benchmark data sets. The distributions are obtained by training the models with 10 different random seeds and recording the obtained DCI score. We use the hyperparameters that \citet{challengingdis} provide in their experiments for the baseline models and share them across all data sets to enable a fair comparison. We observe that our model outperforms all of the considered baselines across all data sets. The used models are: 1) DGP-VAE (ours), 2) Ada-GVAE, 3) AnnealedVAE, 4) \(\beta\)-VAE, 5) \(\beta\)-TCVAE, 6) FactorVAE, 7) DIP-VAE-I, and 8) DIP-VAE-II.}
    \label{fig:baseline_exp}
\end{figure*}


\paragraph{Disentanglement Metric}
\label{metric}
Since there is no universally accepted definition of disentanglement, many previous approaches define their own metrics, leading to a wide variety of such measures \citep{betavae,factorvae,betatcvae,ridgeway18,dipvae,eastwood2018a}. To not introduce another custom metric to the already crowded field and to avoid the impression of quantifying our results with a biased, tailor-made metric, we decided to use one of the existing methods to measure disentanglement.
To this end, we opt for the the DCI Disentanglement score \citep{eastwood2018a}. \citet{challengingdis} show that all commonly used disentanglement metrics are correlated, justifying the use of one of the existing metrics as a proxy for the measurement of disentanglement.


\paragraph{Experimental Results}

We see in \textbf{Fig.}~\ref{fig:baseline_exp} that our approach outperforms all considered baseline methods on all considered data sets. We demonstrate that our model learns the most disentangled representations from the sequential data in all cases, which shows that GP-VAE models are well suited for the task of learning such representations when we consider data with a clear temporal structure. 
We observe that the weakly-supervised Ada-GVAE model is the  runner-up in terms of its disentanglement in all of our experiments, probably due to the fact that it is the only other model that explicitly exploits potential correlations in time.

While the Ada-GVAE approach outperforms the other (fully unsupervised) models, our model outperforms it on time series data with potentially dense changes of underlying factors. 
This is likely due to the fact that the Ada-GVAE is explicitly designed following the assumption that changes in the underlying factors between consecutive time points are sparse, that is, that only a subset of all factors change between two points.
Our model, on the other hand, does not make any such restrictive assumptions, and just assumes that the underlying factors change smoothly over time, as modeled by the chosen GP kernel.
We argue that our more general assumptions about the structure of sequential data are better aligned with data found in the real world and test this hypothesis in the next experiment.

\subsection{Real Medical Data}
\label{real_med}

Real-world clinical time series data commonly consist of noisy, high-dimensional observations. Clinicians are trained to interpret these data and to find patterns that help them identify the underlying causes for these observations. The available observations give clues, but the real interest lies in the underlying health states that give rise to the data. For example, the state of the heart will influence the observed blood pressure and heart rate over time, while the state of the lungs is reflected in respiratory pressure and oxygen saturation of the blood. Learning disentangled representations from high-dimensional medical time series could allow us to infer the state of independent clinical entities from these data, making the health states more salient.

\begin{figure*}[t]
    \centering
    \includegraphics[width=\textwidth]{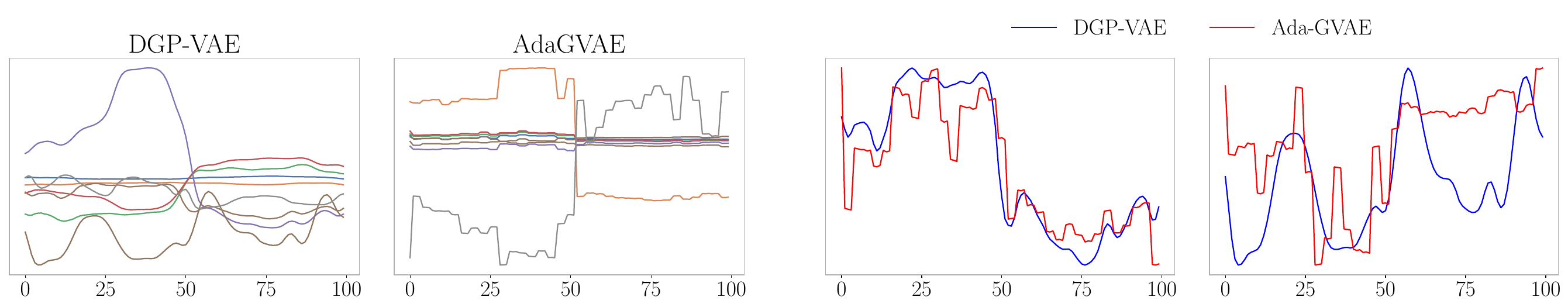}
    \caption{(\textbf{left}) Comparison of the learned latent time series of the same input between our model and the Ada-GVAE. We observe that the latent time series in our model displays dynamics on multiple time scales and multiple channels may change at once, while the latent series of the Ada-GVAE model does not display dense changes of latent factors. (\textbf{right}) Details of two latent channels, for our model and Ada-GVAE, respectively. Our model learns smooth latent time series, while the Ada-GVAE exhibits more noisy dynamics.}
    \label{fig:med_latent_ts}
\end{figure*}

\paragraph{Experimental Setup}

To investigate the ability of our model to disentangle real-world medical time series data, we consider the HiRID data set \citep{hirid}, consisting of 18 clinical variables recorded in five minute intervals for over 33,000 patients. We train our model on these data and evaluate the quality of the learned representations, in terms of disentanglement and completeness, and compare them with the Ada-GVAE model's representations trained on the same data.
We chose to compare our model with the Ada-GVAE since it is the only baseline model for disentanglement that explicitly takes the time series aspect into account.
Moreover, it was the second-best model in our previous experiment and therefore the most serious competitor.

Measuring the disentanglement of learned representations on the HiRID data is not as straightforward as for the benchmark experiments since we do not have access to the underlying factors of generation.
However, as mentioned before, we conjecture that we can still meaningfully learn a representation that can be identified as disentangled.
First, we group the different clinical variables into independent clinical concepts, with the help of a medical expert (for details, see Appendix~\ref{app:map}). Then, we train a classifier to predict the observed variables from the learned latent representation. The classifier returns the importance of each latent factor for predicting a given input variable and, by aggregating the variables into the aforementioned groups, we get a distribution over which latent variable each clinical concept is mapped to.
We calculate the disentanglement and completeness of these distributions according to the DCI disentanglement metric \citep{eastwood2018a} in order to enable a quantitative comparison between our model and the Ada-GVAE.

While the resulting disentanglement and completeness scores do not correspond to those obtained in the setting where we have access to the underlying factors of variation, we argue that they are still useful in measuring the quality of the learned representations. A higher completeness score implies that observations belonging to the same clinical concept are mapped to the same latent variable and a higher disentanglement score indicates that independent clinical concepts are mapped to independent latent variables. This reflects the two main objectives of a well-disentangled representation in this setting: variables with a shared underlying cause are mapped to the same latent factor and independent concepts remain independent in the learned latent space.

As an additional evaluation, in a similar spirit to \citet{gpvae}, we define a downstream mortality prediction task as a proxy measure for the informativeness of the representations. We take the latent representations from each model, train a linear classifier to predict the mortality label of each sample, and report the performance in terms of the area under the receiver-operator-chracteristic curve (AUROC).
Intuitively, the clinical outcome (here: mortality) should be more closely related to the true underlying health factors than to the measurable entities.
A more disentangled representation should thus improve the test performance of a downstream classifier trained on this representation, as has indeed been shown by \citet{adagvae}.

\begin{table}[t]
\caption{Comparison of the DCI disentanglement and completeness scores, as well as downstream classification performance (AUROC) on a mortality prediction task, of the learned representations of the HiRID data set between our model and the Ada-GVAE. We see that our model clearly outperforms the Ada-GVAE in terms of all metrics.}
\vskip 0.15in
\begin{center}
\begin{small}
\begin{sc}
    \begin{tabular}{lcc}
        \toprule
        Metric & Ada-GVAE & DGP-VAE \\
        \midrule
        Disentanglement & 0.133\(\pm\) 0.005 &  \textbf{0.325}\(\pm\) \textbf{0.013}
        \\
        Completeness &  0.185\(\pm\) 0.007 & \textbf{0.358}\(\pm\) \textbf{0.013} \\
        AUROC & 0.684\(\pm\) 0.019 & \textbf{0.769}\(\pm\) \textbf{0.010} \\
        \bottomrule
    \end{tabular}
    \label{tab:med_results}
\end{sc}
\end{small}
\end{center}
\vskip -0.1in
\end{table}

\paragraph{Experimental Results}

\begin{figure}[t]
    \centering
    \begin{subfigure}{0.35\textwidth}
    \centering
    \includegraphics[height=4.6cm]{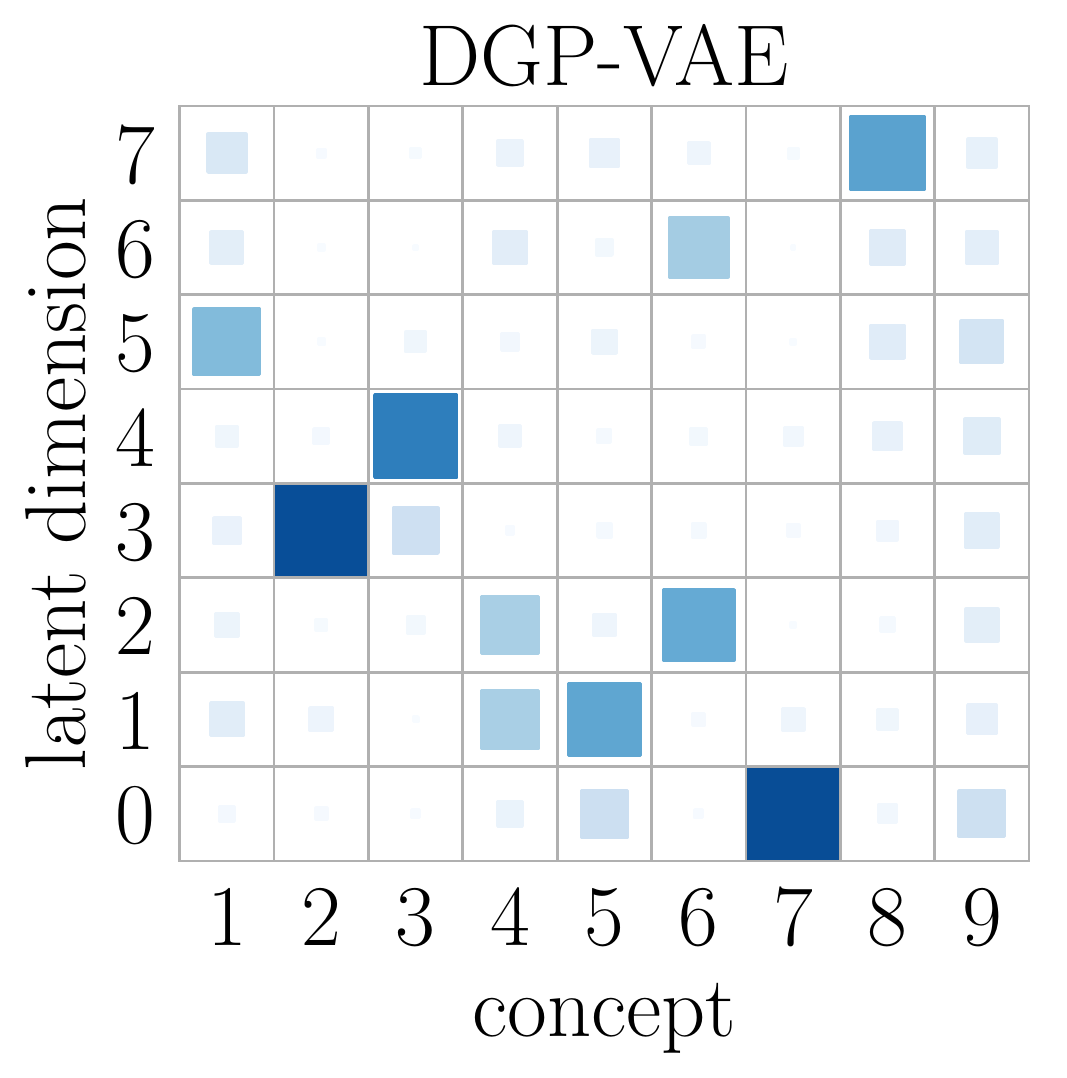}
    \end{subfigure}
    \hspace{0.1cm}
    \begin{subfigure}{0.35\textwidth}
    \centering
    \includegraphics[height=4.6cm]{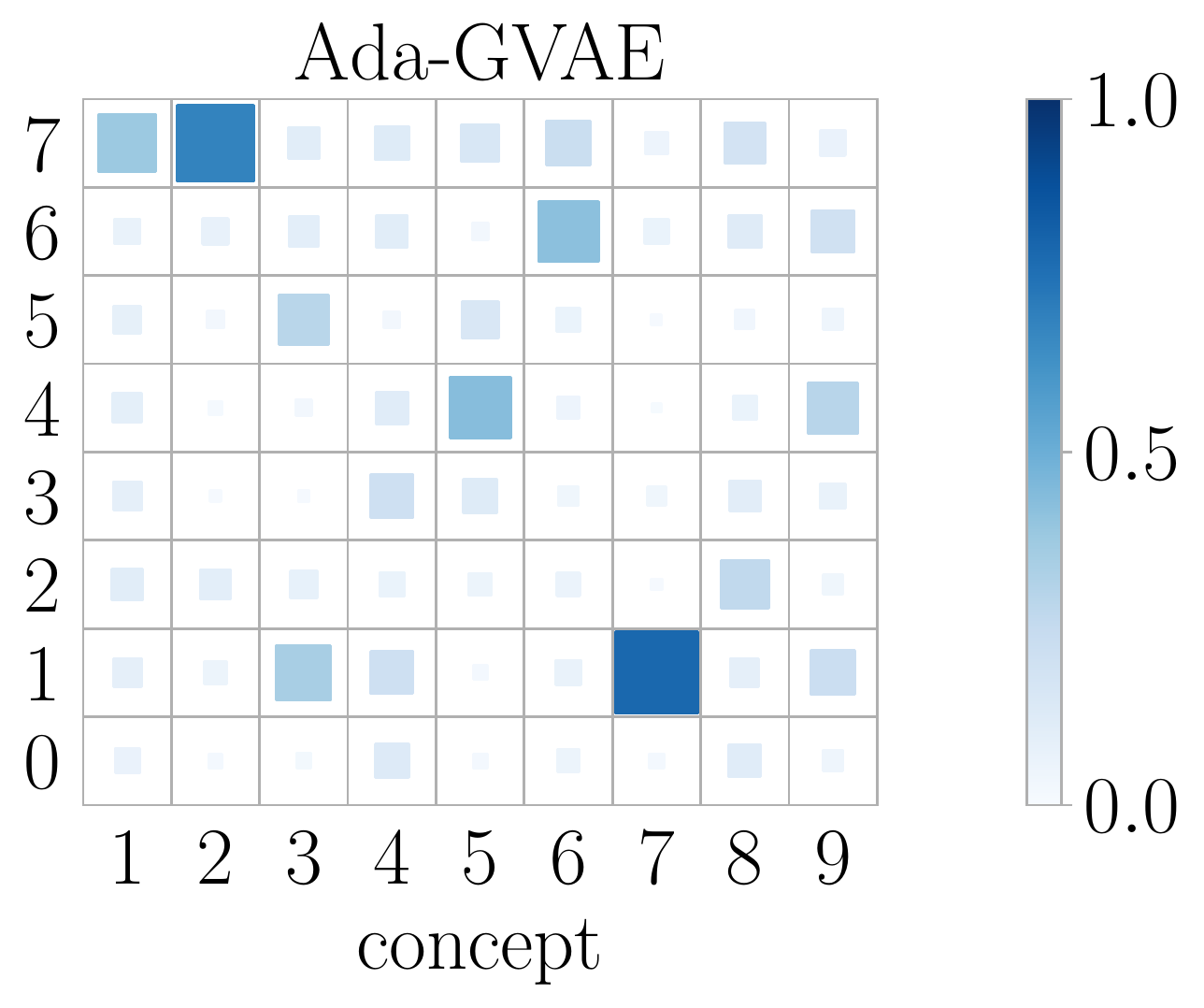}
    \end{subfigure}
    \caption{(\textbf{left}) Mapping of HiRID concepts to latent variables that our model learns. (\textbf{right}) HiRID concept mapping learned by the Ada-GVAE model. We observe that our model learns to cluster related clinical concepts, indicated by features being mapped to a single latent, while disentangling unrelated concepts from each other, that is, mapping independent concepts to independent latents. The Ada-GVAE model learns some of these relations as well, but it is not as successful as our model and also finds spurious relationships in the data. The feature groups are: 1) Cardiovascular, 2) Lungs, 3) Sedation level, 4) Glucose, 5) Blood clotting, 6) Metabolic, 7) Inflammation, 8) Heart medication, and 9) Pain medication. The ground truth mapping of observed HiRID features to clinical concepts is presented in Appendix~\ref{app:map}.}
    \label{fig:med_data}
\end{figure}

We observe that our model outperforms the Ada-GVAE approach both in terms of disentanglement and completeness of the learned representations on the HiRID data set (\textbf{Tab.}~\ref{tab:med_results}). This indicates that we are able to more successfully learn which observed features arise from a shared clinical concept and to separate these independent concepts in the latent space.
Qualitatively, this can also be observed in the comparison of the resulting feature mappings to the latent space, presented in \textbf{Fig.}~\ref{fig:med_data}.
Moreover, our representations also lead to a higher downstream classification performance on the mortality prediction task (\textbf{Tab.}~\ref{tab:med_results}), which further confirms the superiority of our model and highlights the usefulness of its disentangled representations.

The comparison of our model to the Ada-GVAE demonstrates that our assumptions are better aligned with realistically occurring time series data, at least in the medical setting. The experimental results can furthermore be explained by our model's ability to capture long-range dependencies in sequential data, while the Ada-GVAE can only hope to exploit correlations in directly neighbouring samples.
Since we use GP priors to model the latent space, our model also automatically yields a smooth and denoised latent representation time series, possibly providing a more interpretable representation of a patient's physiological state than the original noisy and high-dimensional observations, and also more interpretable than the AdaGVAE, as can be seen in some example visualization in \textbf{Fig.}~\ref{fig:med_latent_ts}.

\section{Conclusion}
\label{conclusion}
In this paper, we investigated the properties of a GP-VAE model to learn disentangled representations from time series data. Our model uses Gaussian process priors to model the latent space together with a structured variational distribution to capture dependencies in time.
We showed that our approach outperforms state-of-the-art models on benchmark disentanglement tasks involving sequential image data. Our modeling assumptions for the structure of time series data are more permissive than those of previous methods \citep{adagvae}, and we provide evidence that they are better aligned with realistically occurring data by showing our model's favorable performance on real-world medical time series data in terms of disentanglement and downstream classification performance.

In the future, it would be interesting to include more specific domain knowledge for certain tasks in the form of specialized GP kernels or hierarchical priors. Furthermore, future work investigating what constitutes a fully disentangled representation seems necessary. In the medical setting, underlying concepts that give rise to observations may be hierarchically related and therefore not fully independent. Currently, this setting cannot be properly captured by the available measures for disentanglement.

\bibliographystyle{abbrvnat}


\newpage
\appendix

\counterwithin{figure}{section}
\counterwithin{table}{section}

\section{Implementation Details}
\label{app:imp_details}

The inference network in our model is implemented with a convolutional neural network (CNN) and for the generative network we use a multilayer perceptron (MLP). Details of the network hyperparameters used in the respective experiments are provided in the following.

\subsection{Benchmark Experiment}
\label{app:bench}
Since all the data sets we consider in our benchmark experiment consist of images, we preprocess them using 2D convolutional layers. This yields a latent representation of the image at each time step, which is then flattened before being used as the input to a 1D convolution over time. We consider subsections of the original time series for the temporal convolution step. The hyperparameters of our model for this experiment are provided in \textbf{Tab.}~\ref{tab:bench_hype}.

All baseline methods that we compare against are implemented with the \texttt{disentanglement\_lib} \citep{challengingdis} and their hyperparameters are given in \textbf{Tab.}~\ref{tab:base_hype}. The evaluation of the DCI metric is also implemented with the \texttt{disentanglement\_lib}. We consider a train/test split of 8000/2000 data points for the classifier used in the DCI score calculation.

\begin{table}[h]
\caption{Hyperparameters used in the DGP-VAE model for the benchmark experiments.}
\vskip 0.15in
\begin{center}
\begin{small}
\begin{sc}
    \begin{tabular}{lr}
        \toprule
        Hyperparameter & Value \\
        \midrule
        Number of CNN layers in inference network & 1 \\
        Number of filters per CNN layer & 32 \\
        Filter size & 3 \\
        Number of feedforward layers in inference network & 2 \\
        Width of feedforward layers & 256 \\
        Dimensionality of latent space & 64 \\
        Length scale of Cauchy kernel & 2.0 \\
        Number of feedforward layers in generative network & 3 \\
        Width of feedforward layers & 256 \\
        Activation function & ReLU \\
        Optimizer & Adam \citep{adam} \\
        Learning rate & 0.001 \\
        Training epochs & 1 \\
        Train/test split (dSprites, SmallNORB) & 10000/500 \\
        Train/test split (Cars3D, Shapes3D) & 6190/310 \\
        Dimensionality if time points (dSprites, SmallNORB) & 4096 \\
        Dimensionality if time points (Cars3D, Shapes3D) & 12288 \\
        Original time series length & 100 \\
        Training time series subsection length & 5 \\
        Tradeoff parameter \(\beta\) & 1.0 \\
        \bottomrule
    \end{tabular}
    \label{tab:bench_hype}
\end{sc}
\end{small}
\end{center}
\vskip -0.1in
\end{table}

\begin{table}[h]
\caption{Hyperparameters used for the baseline models.}
\vskip 0.15in
\begin{center}
\begin{small}
\begin{sc}
    \begin{tabular}{llr}
        \toprule
        Model & Hyperparameter & Value \\
        \midrule
        Ada-GVAE & \(\beta\) & 1.0 \\
        AnnealedVAE & \(c_{max}\) & 25 \\
        & Iteration threshold & 100000 \\
        & \(\gamma\) & 100 \\
        \(\beta\)-VAE & \(\beta\) & 4.0 \\
        \(\beta\)-TCVAE & \(\beta\) & 4.0 \\
        FactorVAE & \(\gamma\) & 30 \\
        DIP-VAE-I & \(\lambda_{od}\) & 5 \\
        & \(\lambda_{d}\) & 10\(\lambda_{od}\) \\
        DIP-VAE-II & \(\lambda_{od}\) & 5 \\
        & \(\lambda_{d}\) & \(\lambda_{od}\) \\
        \bottomrule
    \end{tabular}
    \label{tab:base_hype}
\end{sc}
\end{small}
\end{center}
\vskip -0.1in
\end{table}

\subsection{HiRID Experiment}
\label{app:hirid}
Since the HiRID data does not consist of images, we omit the convolutional preprocessing and perform the 1D convolution time directly on the input data. The details of the utilized hyperparameters for this experiment are provided in \textbf{Tab.}~\ref{tab:hirid_hype}. For the calculation of the DCI score we use a train/test split of 20000/5000.

\begin{table}[h]
\caption{Hyperparameters used in the DGP-VAE model for the HiRID real-world medical data set experiment.}
\vskip 0.15in
\begin{center}
\begin{small}
\begin{sc}
    \begin{tabular}{lr}
        \toprule
        Hyperparameter & Value \\
        \midrule
        Number of CNN layers in inference network & 1 \\
        Number of filters per CNN layer & 128 \\
        Filter size & 12 \\
        Number of feedforward layers in inference network & 1 \\
        Width of feedforward layers & 128 \\
        Dimensionanlity of latent space & 8 \\
        Length scales of Cauchy kernel & [20.0, 10.0, 5.0, 2.5] \\
        Number of feedforward layers in generative network & 2 \\
        Width of feedforward layers & 256 \\
        Activation function & ReLU \\
        Optimizer & Adam \citep{adam} \\
        Learning rate & 0.001 \\
        Training epochs & 1 \\
        Train/test split & 517995/25900 \\
        Dimensionality if time points & 18 \\
        Original time series length & 100 \\
        Training time series subsection length & 25 \\
        Tradeoff parameter \(\beta\) & 1.0 \\
        \bottomrule
    \end{tabular}
    \label{tab:hirid_hype}
\end{sc}
\end{small}
\end{center}
\vskip -0.1in
\end{table}

\section{Additional Results}
\label{app:add_results}

\begin{figure}[h]
\centering
    \includegraphics[width=\textwidth]{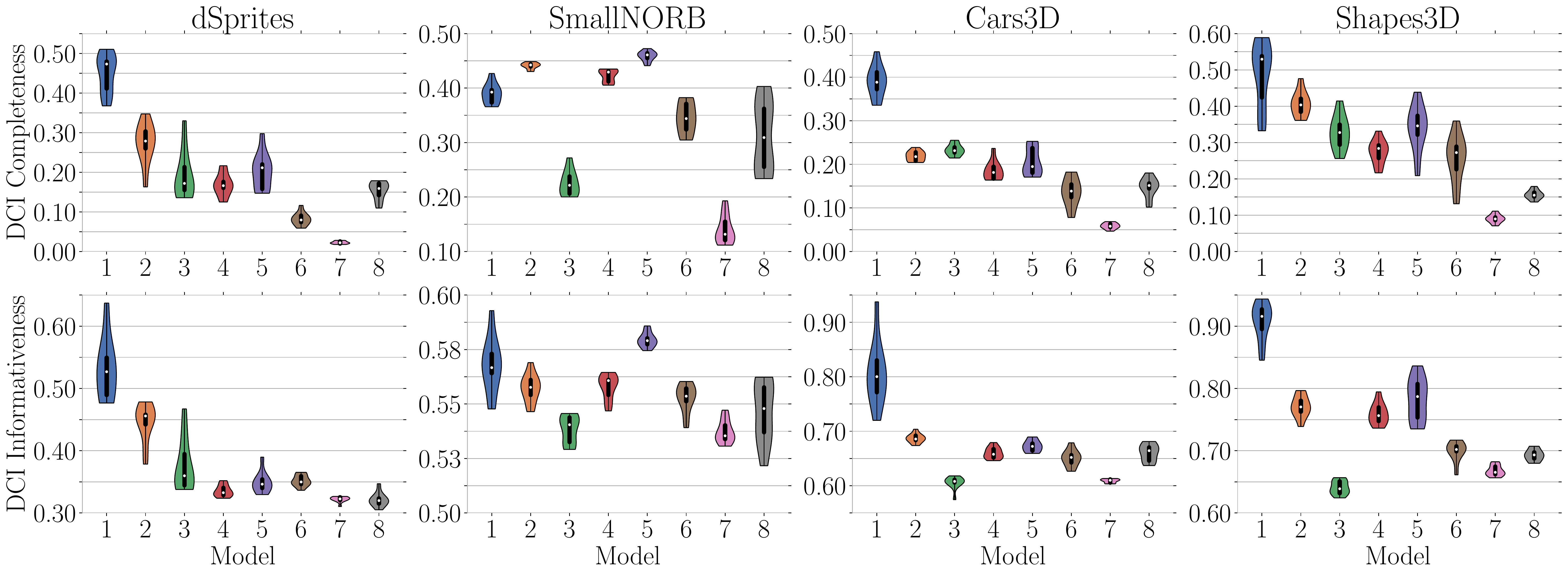}
    \caption{Completeness and informativeness results of the benchmark experiment described in Sec.~\ref{base_methods}. The models we compare are: 1) DGP-VAE (ours), 2) Ada-GVAE, 3) AnnealedVAE, 4) \(\beta\)-VAE, 5) \(\beta\)-TCVAE, 6) FactorVAE, 7) DIP-VAE-I, and 8) DIP-VAE-II.}
    \label{fig:app_dc}
\end{figure}

While we focus on the disentanglement component of the the DCI metric \citep{eastwood2018a}, we also evaluate the completeness and informativeness scores for the benchmark experiment (see \textbf{\textbf{Fig.}}~\ref{fig:app_dc}). Overall, we observe that these additional scores reflect the results reported in Sec.~\ref{base_methods} and that our model outperforms the considered baselines, with the exception of the SmallNORB data set.

\section{Ground Truth HiRID Feature Mapping}
\label{app:map}
To evaluate how well our model can learn disentangled representations of medical time series data we require a ground truth mapping of the occuring medical variables to independent clinical concepts. This mapping, which was provided by a medical expert, is given in \textbf{Tab.}~\ref{tab:hirid_map}.

\begin{table}[h]
\caption{Mapping of HiRID variables into independent clinical concepts.}
\vskip 0.15in
\begin{center}
\begin{small}
\begin{sc}
    \begin{tabular}{llr}
        \toprule
        Concept & Variable & Unit \\
        \midrule
        \textbf{Cardiovascular} & Heart rate & [\(bpm\)] \\
        & Systolic blood pressure (invasive) & [\(mmHg\)] \\
        & Diastolic blood pressure (invasive) & [\(mmHg\)] \\
        & MAP & [\(mmHg\)] \\
        & Cardiac output & [\(l/min\)] \\
        \textbf{Lungs} & SpO2 & [\(\%\)] \\
        & Peak inspiratory pressure (ventilator) & [\(cmH_2O\)] \\
        \textbf{Sedation level} & RASS & [-] \\
        \textbf{Glucose} & Serum glucose & [\(mmol/l\)] \\
        \textbf{Blood clotting} & INR & [-] \\
        \textbf{Metabolic} & Lactate arterial & [\(mmol/l\)] \\
        & Lactate venous & [\(mmol/l\)] \\
        \textbf{Inflammation} & C-reactive protein & [\(mg/l\)] \\
        \textbf{Heart medication} & Dobutamine & Flow [\(mg/min\)] \\
        & Milrinone & Flow [\(mg/min\)] \\
        & Levosimendan & Flow [\(mg/min\)] \\
        & Theophyllin & Flow [\(mg/min\)] \\
        \textbf{Pain medication} & Non-opiod analgesics & [-] \\
        \bottomrule
    \end{tabular}
    \label{tab:hirid_map}
\end{sc}
\end{small}
\end{center}
\vskip -0.1in
\end{table}

\section{Detailed Related Work}
\label{sec:detailed_related}

\subsection{Sequential Data in VAEs}
\label{seq_VAEs}

Models based on variational autoencoders (VAEs) have succesfully been applied to tasks involving---to some extent---the disentanglement of these data. The factorized hierarchical variational autoencoder (FHVAE) introduced by \citet{fhvae} aims to exploit the correlations of sequential data by introducing two different hierarchical priors to the latent representation. The authors argue that this captures the multi-scale nature of sequential data and disentangles features that are shared across a sequence from those that vary from one sequence segment to another.

In a similar spirit, \citet{seqdvae} introduce the Disentangled Sequential Autoencoder, a model that learns to disentangle static from dynamic parts of the data's representation. This is achieved by means of a factorized graphical model that encodes sequence-invariant information into one latent variable and all dynamic information into a separate set of latent variables. While both of the aforementioned models successfully disentangle static from dynamic features, they do not disentangle the underlying, \textit{individual} dynamic factors.

The class of models that employ Gaussian process priors for the latent variables of a variational autoencoder and thereby exploit the correlation of sequential data it the latent space have been successfully applied to a wide range of tasks. The Gaussian Process Prior Variational Autoencoder (GPPVAE) \citep{gppvae} was the first model to introduce this family of priors in the context of VAEs and break with the assumption that samples of the latent distribution must be independent and identically distributed (i.i.d.), thereby better modeling the specifications of sequential data. While the basis of this model is shared with ours, we use one GP per latent channel, as opposed to a joint GP prior over the whole data. This allows us to rely on standard inference techniques as opposed to the specialized inference method of the GPPVAE model, while also encouraging disentanglement between the latent channels.

The Multi-disentangled-features Gaussian Process Variational Autoencoder (MGP-VAE) \citep{mgpvae} extends the GPPVAE model by using fractional Brownian motion and Brownian bridge kernels for the latent GP prior of each channel. The authors argue that this setting allows for the disentanglement of static as well as dynamic features, but only show qualitative results for sparsely changing input time series. In contrast, we show the efficacy of our approach to dynamic sequential data with dense changes of the factors in time.

The Gaussian Process Variational Autoencoder (GP-VAE) model introduced by \citet{gpvae} can be viewed as a further extension to \citet{gppvae}'s GPPVAE model. This model introduces a GP prior with a Cauchy kernel to each latent channel in combination with a structured variational inference technique to impute missing values in time series data. While the GP-VAE was designed and used for missing data imputation we show that a very similar architecture lends itself to learning disentangled representations from dynamic sequential data.

The Sparse GP-VAE (SGP-VAE) \citep{sgpvae} and Scalable GP-VAE (SVGP-VAE) \citep{jazbec2020scalable} extend the class the GP-VAE models with a sparse GP approximation, parameterized by a partial inference network.
Moreover, the Factorized GP-VAE (FGP-VAE) \citep{jazbec2020factorized} further improves the inference speed by using Kronecker-factorized kernels.
These extensions could also be readily applied to our model, which we have not done in this study, since exact inference was still feasible in our experiments.

\subsection{Disentangled Representation Learning}
\label{disent}
All state-of-the-art approaches to disentangled representation learning rely on variational autoencoders \citep{vae} as their architectural backbone. Alternatives based on generative adversarial networks (GANs) \citep{gan} such as variants of InfoGAN \citep{infogan} have also been proposed, but previous work has found their performance to not be competitive when compared to VAE-based approaches \citep{factorvae}.

In the VAE setting, the representation \(r(\textbf{x})\) of a sample \(\textbf{x}\) is taken as the mean of the approximate posterior distribution \(q(\textbf{z}|\textbf{x})\), that is, the encoding of a sample from the feature space to the latent space. The following approaches share the common theme of disentangling this approximate posterior, while their main differences arise from how this disentanglement is enforced.

\paragraph{Unsupervised Models}
\label{unsupervised_bl}
The \(\beta\)-VAE model \citep{betavae} adds a simple hyperparameter to the KL term of the standard ELBO. This \(\beta\) hyperparameter balances reconstruction quality with latent channel capacity and setting it greater than unity enforces the encoder distribution to better match the factorized Gaussian prior. 

The AnnealedVAE model \citep{annealedvae} also focuses on latent bottleneck capacity. The authors argue that limiting the latent channel capacity forces the model to learn a single factor of variation at a time. Therefore, the bottleneck capacity is gradually increased during training to enforce the sequential learning of separate underlying factors of variation.

By further decomposing the vanilla VAE objective, the authors of the \(\beta\)-TCVAE model \citep{betatcvae} identify a term which measures the total correlation between latent variables. By specifically penalizing this total correlation term, the model enforces disentanglement without adding any additional hyperparameters compared to \(\beta\)-VAE.

The FactorVAE model \citep{factorvae} also penalizes the total correlation term, but differs in implementation compared to \(\beta\)-TCVAE. The idea however remains to push the aggregated posterior \(q(\textbf{z})\) towards a factorized form, thus enforcing independence across latent dimensions, without having to sacrifice reconstruction quality for disentanglement.

The authors of the DIP-VAE model \citep{dipvae} propose to enforce disentanglement by means of disentangled priors. The aggregated posterior is pushed towards these disentangled priors by means of penalizing an arbitrary convergence between the two. Subtle implementation differences result in two models: DIP-VAE-I and DIP-VAE-II.

\paragraph{Weakly-Supervised Model}
\label{weak_bl}
The Ada-GVAE model \citep{adagvae} differs from the previously introduced approaches in that it is not a fully unsupervised approach, but utilizes a form of weak supervision to improve disentanglement. The authors acknowledge \citet{challengingdis}'s proof that learning disentangled representations is impossible without inductive biases. They therefore attempt to explicitly include such an inductive bias in their modeling assumptions and exploit this for the purposes of disentanglement. They make the assumption that some underlying factors of variation may be shared across pairs of input data. They then go on to prove that knowing the number of shared factors across individual pairs is sufficient to learn a fully disentangled representation. The introduced model provides an algorithm that is an extension of \(\beta\)-VAE \citep{betavae}, which estimates the number of shared factors in a pair of samples and enforces the sharing of a latent representation for these estimated shared factors. The authors argue that a source of data where this assumption could be justified is sequential data, which inspired us to be explicit about the assumptions we make on sequential data and include these in the form of the inductive bias of smoothly varying GP priors.

\end{document}